\documentclass[11pt, english]{article}
\usepackage{babel}
\usepackage{fullpage,times}
\usepackage{graphicx} 
\usepackage{amsmath,amssymb,amsthm,bm}
\usepackage{times}
\usepackage{subfigure}
\usepackage{natbib}
\usepackage{algorithm}
\usepackage{algorithmic}
\usepackage{hyperref}

\newtheorem{theorem}{Theorem}
\newtheorem{assumption}{Assumption}
\newtheorem{lemma}{Lemma}

\def\zero {{\bf 0}}
\def\a {{\bf a}}

\def\D {{\bf D}}
\def\g {{\bf g}}
\def\p {{\bf p}}
\def\q {{\bf q}}
\def\s {{\bf s}}
\def\x {{\bf x}}
\def\bu {{\bf u}}
\def\bv {{\bf v}}
\def\y {{\bf y}}
\def\z {{\bf z}}
\def\bH {{\bf H}}
\def\I {{\bf I}}
\def\bxi {{\bm \xi}}

\def\prox {{\rm prox}}
\def\tr {{\rm tr}}

\def\E {\mathbb {E}}
\def\R {\mathbb {R}}
\def\sS {\mathcal {S}}
\def\sT {\mathcal {T}}
\def\argmin {\mathop{ {\rm argmin}}}

\title{A Proximal Stochastic Quasi-Newton Algorithm}

\author{
  Luo Luo \\
  Department of Computer Science and Engineering\\
  Shanghai Jiao Tong University, China \\
  \texttt{ricky@sjtu.edu.cn} \\
  \and
  Zihao Chen \\
  Zhiyuan College\\
  Shanghai Jiao Tong University, China \\
  \texttt{zihaochen1996@gmail.com} \\
  \and
  Zhihua Zhang \\
  School of Mathematical Sciences\\
  Peking University, China \\
  \texttt{zhzhang@gmail.com} \\
  \and
  Wu-Jun Li \\
  National Key Laboratory for Novel Software Technology\\ Collaborative Innovation Center of Novel Software Technology and Industrialization \\ Department of Computer Science and Technology\\
  Nanjing University, China \\
  \texttt{liwujun@nju.edu.cn} \\
}

\begin{document}
\maketitle
\date{}

\begin{abstract}
In this paper, we discuss the problem of minimizing the sum of two convex functions: a smooth function plus a non-smooth function. Further, the smooth part can be expressed by the average of a large number of smooth component functions, and the non-smooth part is equipped with a simple proximal mapping. We propose a proximal stochastic second-order method, which is efficient and scalable. It incorporates the Hessian in the smooth part of the function and exploits multistage scheme to reduce the variance of the stochastic gradient. We prove that our method can achieve linear rate of convergence.
\end{abstract}

\section{Introduction}
We consider the following convex optimization problem
\begin{eqnarray}\label{prob:main}
    \min_{\x\in\R^d} P(\x) \overset{\text{def}}= F(\x) + R(\x),
\end{eqnarray}
where $F$ is the average of a set of smooth convex functions $f_i(\x)$, namely
\begin{eqnarray*}
    F(\x) = \frac{1}{n}\sum_{i=1}^n f_i(\x),
\end{eqnarray*}
and $R(\x)$ is convex and can be non-smooth.

The formulation (\ref{prob:main}) includes many applications in machine learning, such as regularized empirical risk minimization. For example, given a training set $\{(\a_1,b_1),(\a_2,b_2),\dots,(\a_m,b_m)\}$, where $\a_i\in\R^d$ is the feature of the $i$th sample and $b_i\in\R$ is the response.
If we take $f_i(\x)=\frac{1}{2}(\a_i^T\x-b_i)^2$, and $R(\x)=\lambda_1||\x||_1$, then we can obtain lasso regression.
If we take $f_i(\x)=\log(1+\exp(-b_i\x^T\a_i))+\lambda_1||\x||_2$ ($b_i\in\{1,-1\}$), $R(\x)=\lambda_2||\x||_1$, then the model becomes logistic regression with elastic net penalty.

One typical approach for solving the formulation (\ref{prob:main}) is first order methods that use proximal mappings to handle the non-smooth part,
such as ISTA \cite{daubechies2003iterative}, SpaRSA \cite{wright2009sparse} and TRIP \cite{kim2010scalable}.
The first order method can be  improved by Nesterov's acceleration strategy \cite{Nesterov:1983}.
One seminal work is the FISTA \cite{beck2009fast}, and related package TFOCS \cite{,becker2011templates} has been widely used.

Another class of methods to handle Problem (\ref{prob:main}) is proximal Newton-type algorithms \cite{fukushima1981generalized,becker2012quasi,oztoprak2012newton,lee2014proximal}.
Proximal Newton-type methods approximate the smooth part with a local quadratic model and successively minimize the surrogate functions.
Compared with the first-order methods, the Newton-type methods obtain rapid convergence rate because they incorporate additional curvature information.

Both conventional first order and Newton-type methods require the computation of full gradient in each iteration, which is very expensive when the number of the component $n$ is very large.
In this case, ones usually exploit the stochastic optimization algorithms, which only process single or mini-batch components of the objective at each step.
The stochastic gradient descent (SGD) \cite{bottou2010large} has been widely used in many machine learning problems.
However,  SGD usually suffers from  large variance of random sampling,  leading to a slower convergence rate.
There are some methods to improve SGD in the case that the objective is smooth (a special case of Problem (\ref{prob:main}) in which $R(\x)\equiv 0$).
They include the first order methods such as SAG \cite{roux2012stochastic} and SVRG \cite{johnson2013accelerating}, and
the Newton-type methods such as stochastic quasi-newton method \cite{byrd2014stochastic}, unified quasi-Newton method \cite{DBLP:conf/icml/Sohl-DicksteinPG14}
and linearly-convergent stochastic L-BFGS \cite{moritz2015linearly}.
There are also some extensions to solve the formulation (\ref{prob:main}) which includes the non-smooth case,
e.g., the first order method Prox-SVRG \cite{xiao2014proximal}, accelerated Prox-SVRG \cite{nitanda2014stochastic} and
proximal stochastic Newton-type gradient descent \cite{shi2015large}.

In this paper, we introduce a stochastic proximal quasi-Newton algorithm to solve the general formulation (\ref{prob:main}).
Our method incorporates the second order information by using a scaled proximal mapping to handle the non-smooth part in the objective.
Compared with \citet{shi2015large}'s stochastic Newton-type method which requires storing the whole data set, our method only needs to store mini-batch data in each iteration. Furthermore,
we exploit the idea of  multistage scheme \cite{johnson2013accelerating,xiao2014proximal} to reduce the variance in our algorithm.
We also prove our method is linearly convergent, which is the same as the special case of solving the smooth problem \cite{moritz2015linearly}.

\section{Notation and Preliminaries}
In this section we give the notation and preliminaries which will be used in this paper.
Let $\I_{p}$ denote the $p \times p$ identity matrix.
For a vector $\a =(a_1, \ldots, a_p)^T \in \R^p$, the Euclidean norm is denoted as $||\a||=\sqrt{\sum_{i=1}^p a_i^2}$ and
the weighted norm is denoted as $||\a||_\bH = \sqrt{\a^T\bH\a}$, where $\bH\in\R^{p\times p}$ is positive definite.
For a subset $\sS\subseteq\{1,2,\dots,n\}$, we define the function $f_\sS$ as
\begin{eqnarray*}
    f_\sS(\x) = \sum_{i\in\sS} f_i(\x).
\end{eqnarray*}
The proximal mapping of a convex function $Q$ at $\x$ is
\begin{eqnarray*}
    \prox_{Q}(\x) = \argmin_{\y}  \; Q(\y) + \frac{1}{2} ||\y-\x||^2.
\end{eqnarray*}
The scaled proximal mapping of the convex function $Q$ at $\x$ with respect to the positive definite matrix $\bH$ is
\begin{eqnarray*}
    \prox_{Q}^{\bH}(\x) = \argmin_{\y} \; Q(\y) + \frac{1}{2} ||\y-\x||_\bH^2.
\end{eqnarray*}

We make the following assumptions.
\begin{assumption}\label{asm:f}
    The component function $f_i$ is $\mu_i$-strongly convex and its gradient is Lipschitz continuous with constant $L_i$; that is, for any $\x,\y\in\R^{d}$, we have
    \begin{eqnarray*}
        \frac{\mu_i}{2}||\x-\y||^2 \leq f_i(\y) - f_i(\x) - (\x-\y)^T\nabla f_i(\y) \leq \frac{L_i}{2}||\x-\y||^2,
    \end{eqnarray*}
    which is equivalent to
    \begin{eqnarray*}
        \mu_i\I_d \preceq \nabla^2 f_i(\x) \preceq L_i\I_d.
    \end{eqnarray*}
\end{assumption}
    Then $F(\x)=\frac{1}{n}\sum_{i=1}^nf_i(\x)$ is $\mu$-strongly convex and its gradient is Lipschitz continuous with constant $L$, where $\mu\geq\frac{1}{n}\sum_{i=1}^n \mu_i$ and $L\leq\frac{1}{n}\sum_{i=1}^n L_i$. Furthermore, let $L_\sS=\sum_{i\in\sS} L_i$.
\begin{assumption}\label{asm:fS}
    For any nonempty size-$b_H$  subset $\sS\subseteq\{1,\dots,n\}$, we have
    \begin{eqnarray*}
        \lambda\I_d \preceq \nabla^2 f_\sS(\x) \preceq \Lambda\I_d.
    \end{eqnarray*}
\end{assumption}
Based on Assumption \ref{asm:f} and the convexity of $R$, we can derive that $P$ is $\mu$-strongly convex
even when $R$ is not strongly convex.

\section{The Proximal Stochastic Quasi-Newton Algorithm}\label{sec:problem}

The traditional proximal Newton-type methods \cite{fukushima1981generalized,becker2012quasi,oztoprak2012newton,lee2014proximal}
use the following update rule at $k$th iteration
\begin{eqnarray}
    \x_{k+1} = \prox_{\eta_k R}^{\bH_k} (\x_k-\eta_k \bH_k^{-1}\nabla F(\x_k)), \label{eq:PQN}
\end{eqnarray}
where $\eta_k$ is the step size and $\bH_k$ is the Hessian of $F$ at $\x_k$ or its approximation.
We can view such iteration as minimizing the composite of local quadratic approximation to $F$ and the non-smooth part $R$,  that is,
\begin{eqnarray*}
    &¡¡&\prox_{\eta_k R}^{\bH_k} (\x_k-\eta_k \bH_k^{-1}\nabla F(\x_k)) \\
    &=& \argmin_{\y} \nabla F(\x_k)^T(\y{-}\x_k) + \frac{1}{2\eta_k}||\y{-}\x_k||_{\bH_k}^2 + R(\y).
\end{eqnarray*}
The update rule (\ref{eq:PQN}) requires the computation of the full gradient $\nabla F(\x_k)$ at each iteration.
When the number of the component $n$ is very large, it is very expensive.
In this case, we can use the stochastic variant of proximal Newton-type methods.
We can sample a mini-batch $\sS_k\subseteq\{1,2,\dots,n\}$ at each stage and take the iteration as follow
\begin{eqnarray}
     \x_{k+1} = \prox_{\eta_k R}^{\bH_k} (\x_k{-}\eta_k \bH_k^{-1}\nabla f_{\sS_k}(\x_k)),
\end{eqnarray}
where $f_{\sS_k}= \sum_{i\in\sS_k}f_i(\x)$.
To avoid the step size $\eta_k$ decaying to zero, we use the multi-stage scheme \cite{johnson2013accelerating,xiao2014proximal}
to reduce the variance in random sampling.
Specifically, we replace $\nabla f_{\sS_k}(\x_k)$ by the variance reduced gradient $\bv_k$:
\begin{eqnarray}
    \bv_k=\frac{1}{Mbq_{\sS_k}}(\nabla f_{\sS_k}(\x_k) - \nabla f_{\sS_k}(\tilde{\x}))  + \nabla F(\tilde{\x}), \label{eq:vk}
\end{eqnarray}
where $b=|\sS_k|$ is the size of mini-batch, $M={n\choose b}$ and $q_{\sS_k}$ is the probability of sampling mini-batch $\sS_k$.
The estimate $\tilde{\x}$ in (\ref{eq:vk}) is the estimate of optimal solution $\x_*$, and we update the full gradient $\nabla F(\tilde{\x})$ after every $m$ iterations.
The probability $q_{\sS_k}$ is proportional to the Lipschitz constant of $\nabla f_{\sS_k}$.  We provide the detailed analysis in Lemma \ref{lemma:variance}.

Thus we use the following modified update rule in our algorithm
\begin{eqnarray}
     \x_{k+1} = \prox_{\eta_k R}^{\bH_k} (\x_k-\eta_k \bH_k^{-1}\bv_k). \label{eq:SPQN}
\end{eqnarray}
If $R$ has simple proximal mapping, the subproblem (\ref{eq:SPQN}) can be solved by iterative methods such as FISTA \cite{beck2009fast}.
When the dimension $d$ is large, solving (\ref{eq:SPQN}) by using the exactly Hessian matrix in each iteration is unacceptable.
To make the iteration (\ref{eq:SPQN}) efficient, we construct the approximation of Hessian by
combining the idea of the stochastic LBFGS \cite{byrd2014stochastic} and the proximal splitting method \cite{becker2012quasi}.
Suppose that the approximate Hessian has the form $\bH_k=\D+\bu\bu^T$, where $\D$ is a diagonal with positive diagonal elements $d_i$ and
$\bu\in\R^d$ is obtained via  the results of recently $2Z$ iterations.
The detail of constructing the Hessian is given in Algorithm \ref{Alg:QuasiHessian}.
We  solve the subproblem (\ref{eq:SPQN}) in terms of the following lemma \cite{becker2012quasi}.
\begin{lemma}\label{lemma:splitting}
    Let $\bH=\D+\bu\bu^T$ be positive definite. Then
    \begin{eqnarray*}
        \prox_Q^\bH(\x)=\D^{-1/2}\prox_{Q\circ\D^{-1/2}}(\D^{1/2}\x-\bv),
    \end{eqnarray*}
    where $\bv=\beta_0\D^{-1/2}\bu$ and $\beta_0$ is the root of
    \small{
    \begin{align*}
        \bu^T\Big(\x-\D^{-1/2}\prox_{Q\circ\D^{-1/2}}\big(\D^{1/2}(\x-\beta\D^{-1}\bu)\big)\Big)+\beta=0.
    \end{align*}}
\end{lemma}
Lemma \ref{lemma:splitting} implies that we can solve the subproblem (\ref{eq:SPQN}) efficiently when the proximal mapping of $R(\x)$ is simple.
We summarize the whole procedure of our method in Algorithm \ref{Alg:ProxSVRG}.
\begin{algorithm}[ht]
    \caption{Proximal Stochastic Quasi-Newton}\label{Alg:ProxSVRG}
    \begin{algorithmic}
    \STATE Initialize $\x_0=\zero$, $r=0$, parameter $m$, $L$, batch size of $b=|\sS|$ and $b_H=|\sT|$ and step size $\eta$
    \STATE \textbf{for} $s=1,2,3\dots$ \textbf{do}
    \STATE \hspace{0.25cm} $\x_0=\tilde{\x}=\tilde{\x}_{s-1}$
    \STATE \hspace{0.25cm} $\tilde{\bv}=\nabla F(\tilde{\x})$
    \STATE \hspace{0.25cm} \textbf{for} $k=1,2,3\dots,m$
    \STATE \hspace{0.50cm}   sample a $b$ size mini-batch $\sS_k\subseteq \{1,\dots,n\}$ \\[0.15cm]
    \STATE \hspace{0.50cm}    $\bv_k=(\nabla f_{\sS_k}(\x_k) - \nabla f_{\sS_k}(\tilde{\x})) / (Mbq_{\sS_k}) + \nabla F(\tilde{\x})$ \\[0.15cm]
    \STATE \hspace{0.50cm}    \textbf{if} $(s-1)m+k < 2Z$ \textbf{then}
    \STATE \hspace{0.75cm}      $\x_{k+1} = \prox_{\eta R} (\x_k-\eta \bv_k)$ \\[0.15cm]
    \STATE \hspace{0.50cm}    \textbf{else}
    \STATE \hspace{0.75cm}      $\x_{k+1} = \prox_{\eta R}^{\bH_r} (\x_k-\eta {\bH_r}^{-1}\bv_k)$ \\[0.15cm]
    \STATE \hspace{0.50cm}    \textbf{end if}
    \STATE \hspace{0.50cm}    \textbf{if} $k \equiv 0\ ({\rm mod}\  Z)$ \textbf{then}
    \STATE \hspace{0.75cm}      $r=r+1$ \\[0.15cm]
    \STATE \hspace{0.75cm}      $\hat{\x}_r=\frac{1}{Z}\sum_{j=k-Z}^{k-1}\x_j$ \\[0.15cm]
    \STATE \hspace{0.75cm}      sample a $b_H$ size mini-batch $\sT_r\subseteq \{1,\dots,n\}$
    \STATE \hspace{0.75cm}      define $\nabla^2 f_{\sT_r}(\hat{\x}_r)$ based on $\sT_r$
    \STATE \hspace{0.75cm}      compute $\s_r=\hat{\x}_r-\hat{\x}_{r-1}$
    \STATE \hspace{0.75cm}      compute $\y_r=\nabla^2 f_{\sT_r}(\hat{\x}_r)\s_r$
    \STATE \hspace{0.75cm}      construct $\bH_r$ as Algorithm \ref{Alg:QuasiHessian}
    \STATE \hspace{0.50cm}    \textbf{end if}
    \STATE \hspace{0.25cm} \textbf{end for}
    \STATE \hspace{0.25cm} $\tilde{\x}_s=\frac{1}{m}\sum_{k=1}^m\x_k$
    \STATE \textbf{end for}
    \end{algorithmic}
\end{algorithm}

\section{Convergence Analysis}\label{sec:exper}

By the strongly convexity of $f_i$, we  show that the eigenvalues of the approximate Hessian $\bH_r$ obtained from Algorithm \ref{Alg:QuasiHessian} is bounded.
\begin{theorem}\label{thm:boundofH}
    By Assumption \ref{asm:fS}, there exist two constants $0\leq\gamma\leq\Gamma$ such that
    the matrix $\bH_r$ constructed from Algorithm \ref{Alg:QuasiHessian}  satisfies
    $\gamma\I_d \preceq \bH_r \preceq \Gamma \I_d$, where
    \begin{eqnarray*}
        \Gamma &=& \frac{d\Lambda}{\alpha}, \\
        \gamma &=& \frac{\alpha(\alpha-2)\lambda^{d+1} + \alpha(1-\alpha)\lambda^d\Lambda + \Lambda^2\lambda^{d-1}}{d^{d-1}\Lambda^d\lambda^2(1-\alpha)}.
    \end{eqnarray*}
    \begin{proof}
        By Assumption \ref{asm:fS} and  Algorithm \ref{Alg:ProxSVRG},
        we have $\lambda\I_d \preceq \nabla^2 f_{\sT_r}(\hat{\x}_r) \preceq \Lambda \I_d$ and $\y_r=\nabla^2 f_{\sT_r}({\hat\x}_r)\s_r$,
        which implies
        \begin{equation} \label{eq:sy1}
            \lambda \leq \frac{\s_r^T\y_r}{||\s_r||^2} \leq \frac{\s_r^T\nabla^2 f_{\sT_r}(\hat{\x}_r)\s_r}{||\s_r||^2} \leq \Lambda.
        \end{equation}
        Letting $\z_r=(\nabla^2 f_{\sT_r}(\hat{\x}_r))^{1/2}\s_r$ and using the definition of $\tau$ in Algorithm \ref{Alg:QuasiHessian}, we have
        \begin{eqnarray}
            \frac{1}{\Lambda} \leq \tau = \frac{\s_r^T\y_r}{||\y_r||^2}
                = \frac{\s_r^T\nabla^2 f_{\sT_r}(\hat{\x}_r)\s_r} {\s_r^T(\nabla^2 f_{\sT_r}(\hat{\x}_r))^2\s_r}
                = \frac{\z_r^T\z_r} {\z_r^T\nabla^2 f_{\sT_r}(\hat{\x}_r)\z_r} \leq \frac{1}{\lambda} \label{eq:sy2}.
        \end{eqnarray}
        Together with (\ref{eq:sy1}) and (\ref{eq:sy2}), we have
        \begin{eqnarray}
            \frac{1}{\Lambda^2} \leq \frac{||\s_r||^2}{||\y_r||^2} \leq \frac{1}{\lambda^2}. \label{eq:sy3}
        \end{eqnarray}
        Using the Woodbury formula and the procedure of Algorithm \ref{Alg:QuasiHessian}, we can write $\bH_r$ as
        \begin{eqnarray*}
                \bH_r
            = (\alpha\tau\I_d + \bu_r\bu_r^T)^{-1}
            = \frac{1}{\alpha\tau}\I_d - \frac{\bu_r\bu_r^T}{\alpha\tau(\alpha\tau + \bu_r^T\bu_r)}.
        \end{eqnarray*}
        Then the largest eigenvalue of $\bH_r$ has the upper bound
        \begin{eqnarray*}
                \sigma_{\max}(\bH_r)
            &\leq& \tr(\bH_r) \\
            &=& \frac{1}{\alpha\tau} \tr\Big(\I_d\Big) - \tr\Big(\frac{\bu_r\bu_r^T}{\alpha\tau(\alpha\tau + \bu_r^T\bu_r)}\Big) \\
            &\leq&  \frac{1}{\alpha\tau} \tr\Big(\I_d \Big)
            = \frac{d}{\alpha\tau}
            \leq \frac{d\Lambda}{\alpha}.
        \end{eqnarray*}
        Then we can bound the value of $\bu_r^T\bu_r$ as follows
        \begin{eqnarray}
                \bu_r^T\bu_r
            &=& \frac{||\s_r-\alpha\tau\y_r||^2}{(\s_r-\alpha\tau\y_r)^T\y_r} \nonumber \\
            &=& \frac{||\s_r||^2 - 2\alpha\tau\s_r^T\y_r + \alpha^2\tau^2||\y_r||^2}{\s_r^T\y_r - \alpha\tau||\y_r||^2} \nonumber\\
            &=& \frac{||\s_r||^2 - 2\alpha\tau^2||\y_r||^2 + \alpha^2\tau^2||\y_r||^2}{\tau||\y_r||^2 - \alpha\tau||\y_r||^2} \nonumber\\
            &=& \frac{||\s_r||^2 - \alpha(2-\alpha)\tau^2||\y_r||^2}{\tau(1 - \alpha)||\y_r||^2} \nonumber\\
            &=& \frac{||\s_r||^2}{\tau(1 - \alpha)||\y_r||^2}  - \frac{\alpha(2-\alpha)\tau}{1 - \alpha} \nonumber\\
            &\leq& \frac{\Lambda}{\lambda^2(1 - \alpha)}  - \frac{\alpha(2-\alpha)}{(1 - \alpha)\Lambda}, \label{eq:uu}
        \end{eqnarray}
        where the last inequality uses the result of (\ref{eq:sy3}).
        We can compute the determinant of $\bH_r$ as follows.
        {\small
        \begin{eqnarray*}
            \det(\bH_r)
            &=& \det\Big(\frac{1}{\alpha\tau}\I_d - \frac{\bu_r\bu_r^T}{\alpha\tau(\alpha\tau + \bu_r^T\bu_r)}\Big) \\
            &=& \frac{1}{(\alpha\tau)^d}\det\Big(\I_d - \frac{\bu_r\bu_r^T}{\alpha\tau + \bu_r^T\bu_r}\Big) \\
            &=& \frac{1}{(\alpha\tau)^{d-1}(\alpha\tau + \bu_r^T\bu_r)} \\
            &\geq& \Big(\frac{\lambda}{\alpha}\Big)^{d-1} \frac{1}{\frac{\alpha}{\lambda} + \frac{\Lambda}{\lambda^2(1 - \alpha)}  - \frac{\alpha(2-\alpha)}{(1 - \alpha)\Lambda}} \\
            &=& \Big(\frac{\lambda}{\alpha}\Big)^{d-1} \frac{\alpha(\alpha-2)\lambda^2+\alpha(1-\alpha)\Lambda\lambda+\Lambda^2}{\lambda^2\Lambda(1-\alpha)} \\
            &=& \frac{\alpha(\alpha-2)\lambda^{d+1} + \alpha(1-\alpha)\Lambda\lambda^d + \Lambda^2\lambda^{d-1}}{\alpha^{d-1}\Lambda\lambda^2(1-\alpha)} .
        \end{eqnarray*}}
        Combining with the result in (\ref{eq:uu}), we have
        \begin{eqnarray*}
            \sigma_{\min}(\bH_r) &\geq& \frac{\det(\bH_r)}{\sigma_{\max}(\bH_r)^{d-1}} \\
            &=& \frac{\alpha(\alpha-2)\lambda^{d+1} + \alpha(1-\alpha)\lambda^d\Lambda + \Lambda^2\lambda^{d-1}}{\alpha^{d-1}\Lambda\lambda^2(1-\alpha)}\frac{\alpha^{d-1}}{(d\Lambda)^{d-1}} \\
            &=& \frac{\alpha(\alpha-2)\lambda^{d+1} + \alpha(1-\alpha)\lambda^d\Lambda + \Lambda^2\lambda^{d-1}}{d^{d-1}\Lambda^d\lambda^2(1-\alpha)} .
        \end{eqnarray*}
    \end{proof}
\end{theorem}

\begin{algorithm}[ht]
    \caption{Construct the inverse of the Hessian}\label{Alg:QuasiHessian}
    \begin{algorithmic}
    \STATE Given $0<\alpha<1$, $\s_r$ and $\y_r$ \\[0.15cm]
    \STATE \hspace{0.25cm} $\displaystyle \tau = \frac{\s_r^T\y_r}{||\y_r||^2} $  \\[0.15cm]
    \STATE \hspace{0.25cm} \textbf{if} $(\s_r-\alpha\tau\y_r)^T\y_r \leq \epsilon ||\y_r|| \ ||\s_r-\tau\y_r||$ \textbf{then}
    \STATE \hspace{0.50cm}      $\bu_r=\zero$
    \STATE \hspace{0.25cm} \textbf{else}
    \STATE \hspace{0.50cm}      $\displaystyle \bu_r=\frac{\s_r-\alpha\tau\y_r}{\sqrt{(\s_r-\alpha\tau\y_r)^T\y_k}}$
    \STATE \hspace{0.25cm}  \textbf{end if}
    \STATE \hspace{0.25cm}  $\bH_r^{-1} = \tau\I + \bu_r\bu_r^T$
    \STATE \textbf{end for}
    \end{algorithmic}
\end{algorithm}

We generalize Lemma 3.6 in \cite{xiao2014proximal}, by integrating the second-order information.
\begin{lemma}\label{lemma:convergence}
    For any $\x,\bv\in\R^d$ and positive definite $\bH\in\R^{d\times d}$,
    let $\x^+ = \prox_{\eta R}^\bH(\x-\eta \bH^{-1} \bv)$, $g=\frac{1}{\eta}(\x-\x^+)$, and $\Delta=\bv-\nabla F(\x)$. Then we have
    \begin{eqnarray*}
        P(\y)&\geq& P(\x^+) + \g^T\bH(\y-\x) + \Delta^T(\x^+ - \y)
             + (\eta ||\g||_{\bH}^2- \frac{L\eta^2}{2}||\g||^2).
    \end{eqnarray*}
\end{lemma}

Similar with the standard proximal mapping, the scaled proximal mapping also has the non-expansive property \cite{lee2014proximal}.
\begin{lemma}\label{lemma:nonexpansive}
    Suppose $Q$ is a convex function from $\R^d$ to $\R$ and $\bH$ satisfies
    $\gamma\I_d \preceq \bH \preceq \Gamma\I_d$. Let $\p=\prox_Q^\bH(\x)$ and $\q=\prox_Q^\bH(\y)$. Then $||\p-\q|| \leq \frac{\Gamma}{\gamma}||\x-\y||$.
\end{lemma}

We can bound the variance of the stochastic gradient $\bv_k$ as following lemma.
\begin{lemma}\label{lemma:variance}
    Let $\bv_k$ be the definition of (\ref{eq:vk}) and let $q_\sS=L_s/(\sum_{|\sT|=b}L_\sT)$ and $L_Q=\frac{1}{n}\sum_{i=1}^n L_i$. Then we have
    {\small
    \[
      \E ||\bv_k-\nabla F(\x_k)||^2 \leq 4L_Q[P(\x_k)-P(\x_*) {+} P(\tilde{\x}) {-} P(\x_*)].
    \]}
\end{lemma}

Based on the above results, we can obtain the following convergence result of our method.
\begin{theorem}\label{thm:convergence}
    Let $0<\eta<\frac{\gamma^2}{8\Gamma L_Q}$, $\x_*=\argmin_{\x}P(\x)$ and $L_Q$ be the definition of Lemma \ref{lemma:variance}.
    When $m$ is sufficiently large, we have
    \begin{eqnarray*}
        \E[P(\tilde{\x}_s)-P(\x_*)] \leq \rho^s (P(\tilde{\x}_0)-P(\x_*)),
    \end{eqnarray*}
    where
    \begin{eqnarray*}
        \rho = \frac{\Gamma\gamma^2 + 4\eta^2\mu\Gamma L_Q(m+1)}
                    {(\eta\gamma^2 - 4\eta^2\Gamma L_Q)\mu m}
               < 1.
    \end{eqnarray*}
    \begin{proof}
        Applying Lemma \ref{lemma:convergence} with $\x=\x_k$, $\x^+=\x_{k+1}$, $\bv=\bv_k$, $\g=\g_k$,
        $\Delta_k=\bv_k-\nabla F(\x_k)$, $\y=\x_*$ and $\bH=\bH_r$, we have
        \begin{align}
            P(\x_*) \geq P(\x_{k+1}) +  \g_k^T\bH_r(\x_* {-} \x_k) + \Delta_k^T(\x_{k+1} {-} \x_*)
              + (\eta ||\g_k||_{\bH_r}^2- \frac{L\eta^2}{2}||\g_k||^2) \label{eq:applemma2}
        \end{align}
        and $\x_k- \x_{k+1} = \eta \bH_r^{-1}(\bv_k+\bxi_k) = \eta \g_k$.
        Then consider the difference of $\x_*$ and iteration results with respect to $\bH_r$
        \begin{eqnarray}
            & & \|\x_{k+1}-\x_*\|_{\bH_r}^2  \nonumber\\
            &=& ||\x_k - \x_* + \x_{k+1} - \x_k||_{\bH_r}^2 \nonumber\\
            &=& ||\x_k - \x_*||_{\bH_r}^2 + (\x_k - \x_*)^T \bH_r (\x_{k+1} - \x_k)
                 + ||\x_{k+1} - \x_k||_{\bH_r}^2 \nonumber\\
            &=& ||\x_k - \x_*||_{\bH_r}^2 - 2\eta \g_k^T \bH_r (\x_k - \x_*)^T + \eta^2||\g_k||_{\bH_r}^2 \nonumber\\
            &\leq&  ||\x_k - \x_*||_{\bH_r}^2 + 2\eta[P(\x_*) - P(\x_{k+1})]
              - 2\eta\Delta_k^T(\x_{k+1} - \x_*) - (2\eta^2 ||\g_k||_{\bH_r}^2- L\eta^3||\g_k||^2)
              + \eta^2||\g_k||_{\bH_r}^2 \nonumber\\
            &\leq&  ||\x_k - \x_*||_{\bH_r}^2 + 2\eta[P(\x_*) - P(\x_{k+1})]
              - 2\eta\Delta_k^T(\x_{k+1} - \x_*), \label{eq:xkxstar1}
        \end{eqnarray}
        where the first inequality uses the results (\ref{eq:applemma2}) and the second inequality is obtained by $\eta\leq \frac{\gamma^2}{8\Gamma L_Q} \leq \gamma/L$.

        Then we bound $-2\eta\Delta_k^T(\x_{k+1} - \x_*)$. We define the result of proximal mapping of the full gradient as
        \begin{eqnarray}
            \bar{\x}_{k+1} = \prox_{\eta R}^{\bH_r} (\x_k-\eta \bH_r^{-1} \nabla F(\x_k)). \label{eq:proxiter1}
        \end{eqnarray}
        Recall that we obtain $\x_{k+1}$ via
        \begin{eqnarray}
            \x_{k+1} = \prox_{\eta R}^{\bH_r} (\x_k - \eta \bH_r^{-1} \bv_k). \label{eq:proxiter2}
        \end{eqnarray} Then we have
        \begin{eqnarray}
            & &   -2\eta\Delta_k^T(\x_{k+1} - \x_*)   \nonumber\\
            &=& -2\eta\Delta_k^T(\x_{k+1} - \bar{\x}_{k+1} + \bar{\x}_{k+1} - \x_*) \nonumber\\
            &=& -2\eta\Delta_k^T(\x_{k+1} - \bar{\x}_{k+1}) -2\eta \Delta_k^T (\bar{\x}_{k+1} - \x_*) \nonumber\\
            &\leq& 2\eta||\Delta_k|| \ ||\x_{k+1} - \bar{\x}_{k+1}|| - 2\eta \Delta_k^T(\bar{\x}_{k+1} - \x_*) \nonumber\\
            &=& 2\eta||\Delta_k|| \ ||\prox_{\eta R}^{\bH_r}(\x_k - \eta \bH_r^{-1} \bv_k) -
              \prox_{\eta R}^{\bH_r}(\x_k-\eta \bH_r^{-1} \nabla F(\x_k))|| -2\eta \Delta_k^T(\bar{\x}_{k+1} - \x_*)\nonumber \\
            &\leq& \frac{2\eta^2\Gamma}{\gamma}||\Delta_k|| \ ||\x_k - \eta \bH_r^{-1} \bv_k - (\x_k-\eta \bH_r^{-1} \nabla F(\x_k)||
              -2\eta \Delta_k^T(\bar{\x}_{k+1} - \x_*) \nonumber\\
            &=& \frac{2\eta^2\Gamma}{\gamma}||\Delta_k|| \ || \bH_r^{-1} \Delta_k|| - 2\eta \Delta_k^T(\bar{\x}_{k+1} - \x_*) \nonumber\\
            &\leq& \frac{2\eta^2\Gamma}{\gamma^2}||\Delta_k||^2 - 2\eta \Delta_k^T(\bar{\x}_{k+1} - \x_*), \label{eq:xkxstar2}
        \end{eqnarray}
\noindent where the first inequality is obtained by the Cauchy-Schwarz inequality and the second inequality is obtained by
        applying Lemma \ref{lemma:nonexpansive} on the fact (\ref{eq:proxiter1}) and (\ref{eq:proxiter2}).
        We note that $\bar{\x}_{k+1}$ and $\x_*$ are independent of the random variable $\sS_k$ and $\E [\Delta_k]=0$ by fixing $\x_k$. Then
        \begin{eqnarray}
            \E [\Delta_k^T(\bar{\x}_{k+1}-\x_*)] = (\E[\Delta_k])^T(\bar{\x}_{k+1}-\x_*)=0. \label{eq:xkxstar3}
        \end{eqnarray}
        Taking the expectation on (\ref{eq:xkxstar1}) and combine the results of (\ref{eq:xkxstar2}) and (\ref{eq:xkxstar3}), we have
        \begin{eqnarray*}
            & &  \E \|\x_{k+1}-\x_*\|_{\bH_r}^2  \nonumber\\
            &\leq&  \|\x_k - \x_*\|_{\bH_r}^2 + 2\eta\E[P(\x_*) - P(\x_{k+1})]
             - 2\eta\Delta_k^T(\x_{k+1} - \x_*) \nonumber\\
            &\leq&  \|\x_k - \x_*\|_{\bH_r}^2 + 2\eta\E[P(\x_*) - P(\x_{k+1})]
              + \frac{2\eta^2\Gamma}{\gamma^2}\E||\Delta_k||^2 - 2\eta \E[\Delta_k^T(\bar{\x}_{k+1} - \x_*)] \nonumber\\
            &\leq& \|\x_k - \x_*\|_{\bH_r}^2 + 2\eta\E[P(\x_*) - P(\x_{k+1})]
               + \frac{8\eta^2\Gamma L_Q}{\gamma^2} [P(\x_k)-P(\x_*)+P(\tilde{\x})-P(\x_*)].
        \end{eqnarray*}
        Consider $s$ stages, $\tilde{\x}_s = \frac{1}{m}\sum_{k=1}^m \x_k$. Summing over $k=1, 2\dots, m$ on the above inequality and taking the expectation with $\sS_0\dots,\sS_{m-1}$, we have
        \begin{eqnarray*}
            & & \sum_{k=0}^{m-1} \E||\x_{k+1}-\x_*||_{\bH_r}^2   \\
            &\leq& \sum_{k=0}^{m-1} ||\x_k - \x_*||_{\bH_r}^2 + \sum_{k=0}^{m-1} 2\eta\E[P(\x_*) - P(\x_{k+1})]
               + \frac{8\eta^2\Gamma L_Q}{\gamma^2} \sum_{k=0}^{m-1} [P(\x_k)-P(\x_*)+P(\tilde{\x})-P(\x_*)].
        \end{eqnarray*}
        That is
        \begin{eqnarray*}
            & & \E \|\x_{m}-\x_*\|_{\bH_r}^2  \\
            &\leq& \|\x_0 - \x_*\|_{\bH_r}^2 + 2\eta\E[P(\x_*) - P(\x_{m})]
             -(2\eta - \frac{8\eta^2\Gamma L_Q}{\gamma^2}) \sum_{k=1}^{m-1} \E[P(\x_k)-P(\x_*)] \\
            & &    + \frac{8\eta^2\Gamma L_Q}{\gamma^2} [P(\x_0)-P(\x_*)   + m(P(\tilde{\x})-P(\x_*))].
        \end{eqnarray*}
        Since $\tilde{\x}=\x_0$, we have
        \begin{eqnarray*}
            & & \E \|\x_{m}-\x_*\|_{\bH_r}^2 + 2\eta\E[P(\x_{m})-P(\x_*)]
             + (2\eta - \frac{8\eta^2\Gamma L_Q}{\gamma^2}) \sum_{k=1}^{m-1} \E[P(\x_k)-P(\x_*)] \\
            &\leq& \|\x_0 - \x_*\|_{\bH_r}^2 + \frac{8\eta^2\Gamma L_Q(m+1)}{\gamma^2}(P(\tilde{\x})-P(\x_*)).
        \end{eqnarray*}
        Based on the fact $2\eta - \frac{8\eta^2\Gamma L_Q}{\gamma^2} < 2\eta$, we have
        \begin{eqnarray*}
            & & (2\eta - \frac{8\eta^2\Gamma L_Q}{\gamma^2}) \sum_{k=1}^{m} \E[P(\x_k)-P(\x_*)]  \\
            &\leq& \E \|\x_{m}-\x_*\|_{\bH_r}^2 + 2\eta\E[P(\x_{m})-P(\x_*)]
              + (2\eta - \frac{8\eta^2\Gamma L_Q}{\gamma^2}) \sum_{k=1}^{m-1} \E[P(\x_k)-P(\x_*)] \\
            &\leq& \|\x_0 - \x_*\|_{\bH_r}^2 + \frac{8\eta^2\Gamma L_Q(m+1)}{\gamma^2}(P(\tilde{\x})-P(\x_*)).
        \end{eqnarray*}
        By the strongly convexity of $P$ and Theorem \ref{thm:boundofH}, we have $P(\tilde{\x}_s) \leq \frac{1}{m}\sum_{k=1}^m P(\x_k)$
        and $||\tilde{\x}-\x_*||^2_{\bH_r} \leq \frac{2\Gamma}{\mu}||P(\tilde{\x})-P(\x_*)||^2$.
        Then we have
         \begin{eqnarray*}
            & & (2\eta - \frac{8\eta^2\Gamma L_Q}{\gamma^2})m  \E[P(\tilde{\x}_s)-P(\x_*)]   \\
            &\leq& (\frac{2\Gamma}{\mu} + \frac{8\eta^2\Gamma L_Q(m+1)}{\gamma^2}) (P(\tilde{\x}_{s-1})-P(\x_*)).
         \end{eqnarray*}
        Taking
         \begin{eqnarray*}
            \rho = \frac{\frac{2\Gamma}{\mu} + \frac{8\eta^2\Gamma L_Q(m+1)}{\gamma^2}}
                    {(2\eta - \frac{8\eta^2\Gamma L_Q}{\gamma^2})m}
                 = \frac{\Gamma\gamma^2 + 4\eta^2\mu\Gamma L_Q(m+1)}
                    {(\eta\gamma^2 - 4\eta^2\Gamma L_Q)\mu m},
         \end{eqnarray*}
        we obtain the desired result
         \begin{eqnarray*}
            & & \E[P(\tilde{\x}_s)-P(\x_*)] \leq \rho^s (P(\tilde{\x}_0)-P(\x_*)).
         \end{eqnarray*}
    \end{proof}
\end{theorem}

\section{Conclusion}
We propose a stochastic quasi-Newton method to solve the non-smooth strongly convex optimization problem.
With the SVRG-type variance reduction strategy, the algorithm does not require store the gradient of each component.
We also prove the algorithm can achieve linear rate of convergence, which is the same as solving the smooth problem.

\bibliography{draft}
\bibliographystyle{icml2016}

\end{document}